# Enhancement of Image Resolution by Binarization

Aroop Mukherjee
Member IEEE

Soumen Kanrar
Member IEEE

## ABSTARCT
Image segmentation is one of the principal approaches of image processing. The choice of the most appropriate Binarization algorithm for each case proved to be a very interesting procedure itself. In this paper, we have done the comparison study between the various algorithms based on Binarization algorithms and propose a methodologies for the validation of Binarization algorithms. In this work we have developed two novel algorithms to determine threshold values for the pixels value of the gray scale image. The performance estimation of the algorithm utilizes test images with, the evaluation metrics for Binarization of textual and synthetic images. We have achieved better resolution of the image by using the Binarization method of optimum thresholding techniques.

## 1. INTRODUCTION
The Image segmentation is an essential task in the fields of image processing and computer vision. It is a process of partitioning the digital images and is used to locate the boundaries into a finite number of meaning full regions and easier to analyze [8]. The Simplest method for image segmentation is thresholding. Thresholding is an important technique in image segmentation, enhancement and object detection. The output of the thresholding process is a binary image whose gray level value 0 (black) will indicate a pixel belonging to a print, legend, drawing, or target and a gray level value 1 (white) will indicate the background. The main complexity coupled with thresholding in documents applications happen when the associated noise process is non-stationary. The factors that make difficult thresholding action are ambient illumination, variance of gray levels within the object and the background, insufficient contrast, object shape and size non-commensurate with the spectacle.

The lack of objective measures to assess the performance of thresholding algorithms is another handicap. Many methods have been reported in the literature [10, 11, 15, 19, 26, 27, 28]. It can extract the object from the background by grouping the intensity values according to the thresholding value. Thresholding divides the image into patches, and each patch is thresholding by a threshold value that depends on the patch contents. In order to decrease the effects of noise, common practice is to first smooth a boundary prior to partitioning. The Binarization technique is aimed to be used as a primary phase in various manuscript analysis, processing and retrieval tasks. So, the unique manuscript characteristics, like textual properties, graphics, line-drawings and complex mixtures of the layout-semantics should be included in the requirements. On the other hand, the technique should be simple while taking all the document analysis demands into consideration. The threshold evaluation techniques are adapted to textual and non-textual area properties, with the special tolerance and detection to different basic defect types that are usually introduced to images. The outcome of these techniques represents a threshold value proposed for each pixel. These values are used to collect the final outcome of the Binarization by a threshold control module. The approach is to examine the manuscript image surface in order to decide about the Binarization method requirement. The Binarization algorithms are to produce an optimal threshold value for each pixel [28]. Therefore we can verify about the algorithm, is best selected for obtaining the optimum thresholding value. The two different algorithms are then discussed in detail to obtain the optimum threshold values. We have structured the paper in different sections. Section 1 presents the introduction, in section 2 we present the literature review. In the section 3 presents problem description. In the section 4 presents the two algorithms. The result and discussion presents in the section 5. Section 6 presents the conclusion remarks.

## 2. LITERATURE REVIEW
The number of survey was done on thresholding. Lee, Chung and Park [15] conducted a comparative analysis of five global thresholding methods and advanced several useful criteria for thresholding performance evaluation. Weszka and Rosenfeld [12] have defined several evaluation criteria. Palumbo, Swaminathan and Srihari [24] has addressed the issue of document binarization comparing three methods, whereas Trier and Jain [18] had the most extensive comparison basis (19 methods) in the context of character segmentation from complex backgrounds. Sahoo et al. [27] surveyed nine thresholding algorithms and illustrated relatively their performance. Glasbey [5] pointed out the associations and performance difference between 11 histogram-based algorithms based on a wide statistical study. The choice of the most suitable one is not a simple procedure. The assessment and evaluation of these algorithms is difficult to provide evidence, since there is no objective way to compare the results. Leedham et al. [9] compared five binarization algorithms by using the precision and recall analysis of the resultant words in the foreground. He et al. [13] compared six algorithms by evaluating their effect on end-to-end word recognition performance utilizing a commercial OCR engine. Sezgin and Sankur [16] described 40 thresholding algorithms and categorized them according to the used information content. They measured and ranked their performance comparatively in two different contexts of images. The problem is that almost in every case, they try to use results from ensuing tasks in document processing hierarchy, in order to estimate the performance of the binarization algorithm. In case of historical documents where their quality obstructs the recognition, and sometimes the word segmentation as well, this method of evaluation can be proved problematic. On the other hand, we need a different, more direct evaluation technique dealing only with the binarization stage. The ideal way of evaluation should be able to decide, for each pixel, if it has finally succeeded the right color (black or white) after the binarization.

The literature divides thresholding techniques into global and local methods. Global thresholding methods optimize a certain objective function or criterion as it is sensible global to the whole image or its squashed histogram representation. Some of these optimisation criteria include between or among classes variances [17, 25], entropy with many of its variants such as two-dimensional entropy or cross entropy [1,3,4,14,20], correlation between the original image and its thresholded version [16,21,22,23], error minimisation, and maximum likelihood [6,30]. On the other hand, local methods [3, 7], [2, 29] apply their objective function to subimages, or blocks, that make up the whole image. A survey of these methods, local and global, appears in [16]

## 3. PROBLEM DESCRIPTION

Thresholding is a technique used for segmentation, which separates an image into two meaningful regions: foreground and background, through a selected threshold value $T$.
If the image is a grey image, $T$ is a positive real number in the range of $[0,....,K]$, Where, $K \in R^{\geq 0}$. So, thresholding may be viewed as an operation that involves tests against a value $T$ (say). In this work we obtained the optimum value of $T$ by comparing of two novels Binarization algorithms that we have considered.

The segmentation procedure is represented by the following equation:

$$G_B(x, y) = \begin{cases} 1, & \text{if } G(x,y) > T \text{.......(1)} \\ 0, & \text{if } G(x,y) \leq T \text{......(2)} \end{cases}$$

$G(x, y)$ represents the intensity value of pixel at $(x, y)$ location in the grey image $G$, where, $G \in R$. $G_B$ represents the obtained segmentation result such that $G_B \in R^{\geq 0}$.

If $G_B(x, y) = 1$, then pixel location $(x, y)$ in the bimodal image $G$ is classified as a foreground pixel, otherwise it is classified as a background pixel. By comparing Binarization algorithms assign the best optimum threshold value for T.
.

## 4. ALGORITHMS
### 4.1 Procedure

Here we present the first algorithm to find the optimum threshold value of an image

// Store the image in jpg format

**1**.Var
  X : array [1..m,1…n] of real;
**2**. Y: array[1,2] of real;

**3**. Opim: array[1…m,1..n] of real;

**4**. Var
  p, a, im, ht, wd, opim : integer ;

// Initialization
**5**. p = 0,a=0,im=0,ht=0,wd=0;

**6**. BEGIN
// size will return two variable values
// size of image and store in array using
// Find_height_width ()

**7.** Y[ ht ,wd] ← Find_height_width(a)
**8**. opim ← X[ht ,wd]

// Compute mean of array

$$\phantom{a \leftarrow}\ ht$$
**9**. $a \leftarrow \Sigma(im)$
$$\phantom{a \leftarrow}\ i=1$$

**10**. for i = 1 to ht do
**11.**     for j = 1 to wd do
**12**.         p ← X[i,j]
**13**.         if (p < = a) then begin
**14**.             opim(i,j) ← 0;
**15**.         else
**16**.             opim(i,j) ← 255;
**17**.         endif
**28**.     END
**19**. END

$$\phantom{b \leftarrow}\ 255$$
**20**. $b \leftarrow \Sigma(opim)$
$$\phantom{b \leftarrow}\ i=0$$

**21**. print b
**22**. END

### 4.2 Procedure
Here we present the second algorithm to find the optimum threshold value of an image

// store the image in jpg format
**1**. Var
  X: array[1..m,1..n] of real;
**2**. Y: array[1,2] of real;

**3**. Var
  total_mean, total_pixel_number1,
  total_pixel_number2 ,
  pixel_number1, pixel_number2,
  optimum_threshold, total,
  b, m1,m2 estimate_threshold , ht, wd,j ,L :
  integer;

// Initialization
**4**. total_mean=0, total_pixel_number1=0,
  total_pixel_number2=0, pixel_number1=0,
  pixel_number2=0, optimum_threshold=0,
  total=0, b=0, m1=0, m2=0,
  estimate_threshold=0, ht=0, wd=0,
  j=0, L =0;

**5**. BEGIN

$$\phantom{estimate\_threshold \leftarrow}\ N$$
**6**. *estimate_threshold*← $\Sigma(a)$
$$\phantom{estimate\_threshold \leftarrow}\ i=1$$

// size will return two variable values
// size of image and store in array using Find_height_width (
//  )

**7.** Y[ht ,wd] ← Find_height_width(a)
// calling optical () recursive
**8.** optical (m,n,estimate_threshold)

**9.** END

**10.** Sub: optical (ht, wd, estimate_threshold)

**11.** BEGIN
**12.** for j = 1 to estimate_threshold do
**13.**   $pixel$_number1 ← $(j*\Sigma(j))/\Sigma(j)$
**14.**   $total$_pixel_number1 ← $\Sigma pixel$_number1
**15.** END

**16.** for L = estimate_threshold + 1 to wd do
**17.**   $pixel$_number2 ← $(L*\Sigma(L))/\Sigma(L)$
**18.**   $total$_pixel_number2 ← $\Sigma pixel$_number2
**19.** END

// compute mean of array
**20.** $total$_mean ← $\sum_{i=1}^{wd}(m1, m2)$

**21.** if (estimate_threshold − total_mean) < 1
**22.**   optimum_threshold ← total_mean
**23.** else
**24.**   estimate_threshold ← total_mean

// call function recursive
**25.** optical( ht, wd, estimate_threshold)
**26.** ENDIF
**27.** END

## 5. EXPERIMENTAL RESULTS AND DISCUSSION

To verify the operational performance of both the algorithms, a set of various images have been tested by 4.1 & 4.2 methods. For all the tested images, the images labeled (a) are original images and image labeled (b) is the output image by using the Binarization method and the image labeled (c) & (d) represents the histogram of the input and output of the image. The performance of the Binarization algorithms with respect to the mean value of each class is derived. In the Figure 1 the output of the estimated threshold value became 93 and after the optimization, threshold value became 140. In the Figure 2 the output of the estimated threshold value became 93 and after the optimization, threshold value became 75.

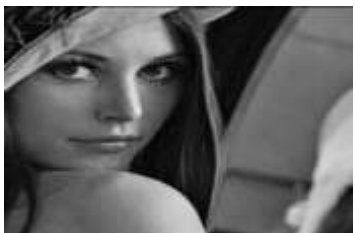

Figure 1(a)

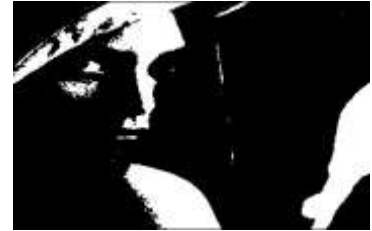

Figure 1(b)

Figure 1(a, b): Using Procedure 1 the estimate threshold is 93 and the optimum threshold value is 140

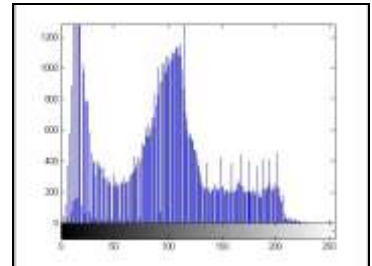

Figure 2(c)

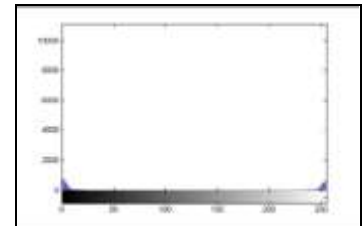

Figure 2(d)

Figure 2(c, d): Using Procedure 1 the histogram on estimate threshold is 93 and the optimum threshold value is 140

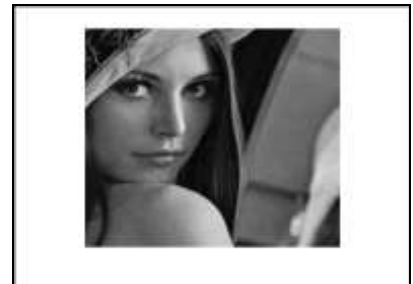

Figure 3(a)

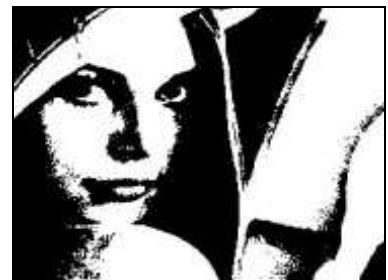

Figure 3(b)

Figure 3(a, b): Using Procedure 2 the estimate threshold is 93 and the optimum threshold value is 75

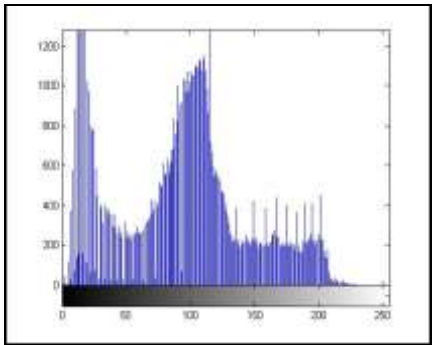

Figure 4(c)

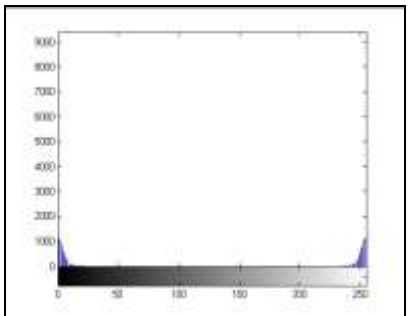

Figure 4(d)

Fig 4(c, d): Using Procedure 2 the histogram on estimate threshold is 93 and the optimum threshold value is 75.

## 6. CONCLUSIONS

This paper proposes a threshold method based on Binarization model. According to the facts, the histogram of an image is used to estimate the mean value of an image. The optimal threshold is determined on the average of the mean value. If the estimated threshold value is 93, the optimum threshold values obtained using 4.1 and 4.2 procedures are 140 and 75 respectively. The optimum threshold value 75 using 4.2 procedures is closer to the estimated threshold value 93. Thus, the results show that 4.2 procedures have achieved a better threshold value compared to the 4.1 procedure.